# Detection of Hate Speech using BERT and Hate Speech Word Embedding with Deep Model

Hind Saleh
Dept. of Computer Science
King Abdulaziz University
University of Tabuk
KSA
h.alatwi@ut.edu.sa

Areej Alhothali
Dept. of Computer Science
King Abdulaziz University
KSA
aalhothali@kau.edu.sa

Kawthar Moria
Dept. of Computer Science
King Abdulaziz University
KSA
kmoria@kau.edu.sa

## Abstract

The enormous amount of data being generated on the web and social media has increased the demand for detecting online hate speech. Detecting hate speech will reduce their neg-ative impact and influence on others. A lot of effort in the Natural Language Processing (NLP) domain aimed to detect hate speech in general or detect specific hate speech such as religion, race, gender, or sexual orientation. Hate communities tend to use abbreviations, intentional spelling mistakes, and coded words in their communication to evade detection, adding more challenges to hate speech detec-tion tasks. Thus, word representation will play an increasingly pivotal role in detecting hate speech. This paper investigates the feasibil-ity of leveraging domain-specific word embed-ding in Bidirectional LSTM based deep model to automatically detect/classify hate speech. Furthermore, we investigate the use of the transfer learning language model (BERT) on hate speech problem as a binary classification task. The experiments showed that domain-specific word embedding with the Bidirec-tional LSTM based deep model achieved a 93% f1-score while BERT achieved up to 96% f1-score on a combined balanced dataset from available hate speech datasets.

## 1 Introduction

Social media has been used extensively for various purposes, such as advertising, business, news, etc. The idea of allowing users to post anything at any time on social media contributed to the existence of inappropriate content on social media. As a result, these platforms become a fertile environment for this type of content. Hate speech is the most com-mon form of destructive content on social media, and it can come in the form of text, photographs, or video. It is defined as an insult directed at a per-son or group based on characteristics such as color, gender, race, sexual orientation, origin, national-ity, religion, or other characteristics (Weber, 2009). Hate speech poses a significant threat to commu-nities, either by instilling hatred in young people against others or by instigating criminal activity or violence against others.

Hate speech on the internet is on the rise around the world, with approximately 60% of the global population (4:54 billion) using social media to com-municate (Ltd, 2020). According to studies, ap-proximately 53 percent of Americans have encoun-tered online harassment and hatred (League, 2019). This score is 12 points higher than the findings of a similar survey performed in 2017 (Duggan, 2017). According to Clement (2019), 21% of students frequently encounter hate speech on social media. Thus, the detection of hate content on social media is an essential and necessary requirement for so-cial media platforms. Social media providers work hard to get rid of this content for a safer social en-vironment. Detecting hateful content is considered one of the challenging NLP tasks as the content might target/attack individuals or groups based on various characteristics using different hate terms and phrases (Badjatiya et al., 2017).

Social media users often employ abbreviations and ordinary words(not hateful) to express their hate intent implicitly that known as code words to evade from being detected (e.g., using Google to refer to dark skin people), which adds extra diffi-culties to detect hate speech. Many studies have proposed machine learning models to handle this problem by utilizing a wide range of features set and machine learning algorithms for classification (Magu et al., 2017; Agarwal and Sureka, 2015; Jaki and De Smedt, 2019; Hartung et al., 2017). These methods often utilize features that require considerable effort and time to be extracted, such as text-based, profile-based, and community-based features. Other studies have worked on linguistic-based features (e.g., word frequency) and deep learning for classification (de Gibert et al., 2018), or distributional based features (e.g., word embed-ding) and machine learning classifier (Gupta and Waseem, 2017; Badjatiya et al., 2017; Djuric et al., 2015; Nobata et al., 2016).

Studies show that distributional features provide a promising result in NLP tasks such as sentiment analysis (Gupta and Waseem, 2017). Recently, deep learning methods also show that it performs well on various NLP problems (Socher et al., 2012). Thus, this study investigates the performance of employing these two methods. Accordingly, this study uses the distributional-based learning method to extract meaningful domain-specific embedding as features and deep learning based on Bidirectional Long Short Term Memory (BiLSTM) classifier to detect hate speech. The word embedding/ distributional representation in this research is built upon a hate speech corpus of 1; 048; 563 sentences to reach the closest meaningful representation vec-tor of hate words. Then, compare it with the domain-agnostic embedding model such as Google Word2Vec and GloVe under the same classifier. We also assess the performance of detecting hate speech using Google's pre-trained BERT model, which has generally achieved a state-of-the-art for many NLP tasks. The contributions of this research are highlighted as follow:

- An unsupervised domain-specific word em-bedding model was developed to extract the meaning of commonly used terminology, acronyms, and purposefully misspelled hate words.

- A comparison between the domain-specific and domain agnostic embedding was provided. The findings show that domain agnostic em-bedding performs slightly better (about 1%), despite the huge difference in the trained cor-pus size.

- The evaluation of a BiLSTM-based deep model with domain-specific embeddings shows an improvement ranging from 5 to 6 points on available datasets over the state-of-the-art techniques.

- The evaluation of the BERT language model on the hate speech binary classification task shows an improvement of about 2 points com-pared to the domain-specific word embedding model.

The remaining of this paper is constructed as follows: the background section provides background information about hate detection and the used methodologies; the review of literature sec-tion includes most recent studies in the field; the methodology section describes the methods used in this study and its specification; the experiment and result section presents the used datasets, embedding models, and results of the experiments; the discussion section includes analysis and obser-vation from the results, and finally the conclusion section summarizes all the findings.

## 2 Background

This section gives an overview of hate speech detec-tion in the field and it provides information about the used methodologies for both of the features and classifiers.

### 2.1 Hate Speech Detection

Several research have attempted to solve the problem of detecting hate speech in general by differentiating hate and non-hate speech. (Ribeiro et al., 2017; Djuric et al., 2015). Others have tackled the issue of recognizing certain types of hate speech, such as anti-religious hate speech. (Albadi et al., 2018; Zhang et al., 2018), jihadist (De Smedt et al., 2018; Ferrara et al., 2016; Wei et al., 2016; Gialam-poukidis et al., 2017), sexist and racist (Badjatiya et al., 2017; Pitsilis et al., 2018; Gamback¨ and Sik-dar, 2017). Several platforms have been used to collect datasets from various online resources such as websites or online forums (e.g., 4Chan, DailyStorm), or recent social media platforms (e.g., Twitter, Facebook). Hate speech has been applied also on different languages (e.g., English, Arabic, German).

### 2.2 Word Embedding

Word embedding (Bengio et al., 2003) is a promi-nent natural language processing (NLP) technique that seeks to convey the semantic meaning of a word. It provides a useful numerical description of the term based on its context. The words are repre-sented by a dense vector that can be used in estimat-ing the similarities between the words (Liu, 2018). The word is represented by an N-dimensional vec-tor appropriate to represent the word meaning in a specific language (Mikolov et al., 2013). The word embedding has been widely used in many recent NLP tasks due to its efficiency such as text classifi-cation (Gamback¨ and Sikdar, 2017; Lilleberg et al., 2015), document clustering (Ailem et al., 2017), part of speech tagging (Wang et al., 2015), named entity recognition (Siencnikˇ, 2015), sentiment anal-ysis (Tang et al., 2014; Wang et al., 2016; Al-Azani and El-Alfy, 2017), and many other problems. The most common pretrained word embedding models are Google Word2Vec, Stanford GloVe and they are described as follow:

### 2.2.1 Word2Vec

Word2Vec is one of the most recently used word embedding models. It is provided by the Google research team (Mikolov et al., 2013). Word2Vec associates each word with a vector-based on its surrounding context from a large corpus. The train-ing process for extracting the word vector has two types, the continuous bag of words model (CBOW), which predicts the target word from its context, and the Skip-Gram model (SG), which predicts the tar-get context from a given word. The feature vector of the word is manipulated and updated accord-ing to each context the word appears in the cor-pus. If the word embedding is trained well, similar words appear close to each other in the dimensional space. The word similarities between the words are measured by the cosine distance between their vec-tors. Google released a vector model called Google Word2Vec that has been trained on a massive cor-pus of over 100 billion words.

### 2.2.2 GloVe

Pennington et al. (2014) provides another popular word embedding model named GloVe (Global Vec-tors for Word Representation). GloVe learns em-beddings using an unsupervised learning algorithm that is trained on a corpus to create the distribu-tional feature vectors. During the learning process, a statistics-based matrix is built to represent the words to words co-occurrence of the corpus. This matrix represents the word vectors. The learning process requires time and space for the matrix con-struction, which is a highly costly process. The difference between GloVe and Word2Vec is in the learning process, Word2Vec is a prediction based model, and GloVe is a count-based model. The GloVe is learned from Wikipedia, web data, Twit-ter, and each model is available with multiple vector dimensions.

### 2.3 Bidirectional Long Short-Term Memory (BiLSTM)

LSTM (Hochreiter and Schmidhuber, 1997) is an enhanced version of the recurrent neural network, which is one of the deep learning models that is designed to capture information from a sequence of information. It differs from the feed-forward neural network in that it has a backward connection. RNN suffers from a vanishing gradient problem that happens when the weights are not updated anymore due to the small value of the received from error function in respect to the current weights in the iteration. The value is vanishing in very long sequences and becomes close to zero. This problem stops RNN from training. LSTM solves this problem by adding an extra interaction cell to preserve long sequence dependencies. Thus, LSTM saves data for long sequences, but it saves the data only from left to right. However, to save sequence data from both directions, a Bidirectional LSTM (BiLSTM) is used. BiLSTM consist of two LSTM, one process the data from left to right and the other in opposite direction then concatenates and flattens both forward and backward LSTM to improve the knowledge of the surrounding context.

### 2.4 BERT Pre-trained Language Model

Bidirectional Encoder Representations from Trans-formers (BERT) (Devlin et al., 2018) is a language model trained on very huge data based on con-textual representations. BERT consists of feature extraction layers, which consist of word embed-ding and layer for the model (e.g., Classification, Question Answering, Named Entity Recognition). BERT is the most recent language model and pro-vides state of the art results in comparison to other language models for various NLP tasks. BERT training procedure of word embedding differs from other word embedding models. It creates a bidirec-tional representation of words that may be learned from both left and right directions. Word embed-ding approaches like Word2Vec and GloVe only examine one direction (either left to right or right to left), resulting in static word representations that do not alter with context. If the word's meaning varies depending on the context, GloVe and Word2Vec map the word to only one embedding vector. As a result, Word2Vec and GloVe are referred to as context-free models. BERT is also different from previous language models (e.g., ELMo stands for Embeddings from Language Models (Peters et al.,

2018)) in that it manipulates the context in all layers in both directions (left and right). Instead of shal-low combining processes such as concatenating, it use cooperatively conditioning to combine both left and right context. BERT is trained on Books Cor-pus (800M words) and English Wikipedia (2,500M words) (Devlin et al., 2018).

## 3 Review of Literature

It is worth noting that word embedding is an effective approach to a variety of NLP issues. To extract bio-events from the scientific literature, Li et al. (2015) used word embedding. They used multiple sets of features such as, word embedding, BOW + n-gram joint model, and word embedding BOW joint model with SVM classifier and the overall performance of word embedding BOW is better than other models on different events, which reached to 77:37% f1-score. The pure word embedding model has lower performance because the dataset size is small. Wu et al. (2015) also used word embedding to distinguish clinical abbreviations as a special case of word sense disambiguation (WSD). The performance of SVM increased when employing word embedding features with an average accuracy of 93%.

Hate speech identification is a prevalent issue that has gotten a lot of attention from researchers. Liu (2018) employed domain-specific word embedding model trained on the articles from hate speech websites and high centrality users' tweets to reach to the semantics of code words used in hate speech. They experimented on CNN, and LSTM models and concluded that CNN performed better than LSTM on tweets due to the length of tweets. The achieved f1-score is 78% given that they experimented on the previous tweet-length 180 characters. Gupta and Waseem (2017) evaluate the performance of using hate Word2Vec (i.e., domain-specific) model with Logistic Regression (LR) classifier on three datasets. They achieved up to 91% f1-score. The results showed that domain-specific word embedding has a desirable performance and is suitable for unbalanced classes datasets.

Nobata et al. (2016) aimed to detect abusive language using pretrained word embeddings on two domains (finance and news) and regression model for classification, they achieved 60:2% and 64:9% f1-score respectively. The results showed that Google Word2Vec provides better performance with 5% on both domains. Badjatiya et al. (2017) employed deep learning techniques to extract embedding features from hate speech text and then used a decision tree model for classification. They reached 93% f1-score using random embeddings initialization that is fed to LSTM to construct features. The results proved that domain-specific embedding can provide better representation of hate words such as "racist" or "sexist" words, because it can extract the meaning of frequently used terms by the hate community, domain-specific-based detection is a promising method for the detection of hate speech, according to all of the research above.

The authors of (Devlin et al., 2018) looked into the BERT model's performance on a variety of NLP tasks. On eleven of these tasks, the model produced state-of-the-art results. It improved by 7:7 points in the General Language Understanding Evaluation (GLUE) benchmark, 4:6 points in Multi-Genre Nat-ural Language Inference (MultiNLI), and 1:5 to 5:1 points in the SQuAD various versions question an-swering tests.

BERT language model was recently employed in a shared task to detect offensive language (Zhu et al., 2019; Pelicon et al., 2019; Wu et al., 2019). Zhu et al. (2019) fine tuned BERT model for this task and came in third place among competitors. They used 13; 240 in tweets to train the algorithm, with each message categorized as offensive or not offensive. They achieved 83:88% f1-score. Mozafari et al. (2020) studied the performance of the BERT language model on hate speech detection as a multi-class problem in a recently released work. They employed a BERT basis and a variety of classifiers, including CNN, which provided the highest score that reached to 92% f1-score.

## 4 Methods

This section describes the used methodologies to handle the detection of hate speech. Mainly, there are two approaches used in this study seeking to find the best classification performance.

- Approach 1: Domain-specific embedding features with BiLSTM based deep model classifier

The reason for using domain-specific (hate domain) word embedding is to construct the vector that represents a closer meaning of hate terms and abbreviations (e.g., black). The classifier used in this approach is a deep model constructed from Bidirectional LSTM to preserve long dependencies

of the input text from both directions. The following steps describe the first approach in details:

Data Collection: The goal of this step is to build a large data corpus to be utilized for embed-ding extraction. The data collected consists of 1; 048; 563 sentences from available hate speech datasets (Golbeck et al., 2017), (Founta et al., 2018), (Davidson et al., 2017), (Waseem and Hovy, 2016), and (Waseem, 2016), in addition to a dataset we collected in this study from Twitter using commonly known hate keywords (e.g., n*gga, f*ck, s*ave) from a pre-defined lexicon includes hate speech words and expressions, called HateBase (Inc, 2020), also from accounts who have an explicit hate content in their tweets or user names (e.g., _@Na***e and Race), or share hate words or phrases in hate hashtags.

Pre-processing: The pre-processing stage was performed to remove any non-meaningful words and symbols such as (stop words, punctuation), stop words are excluded because the presence or absence of these words is not important to extract the meaning of the word in case we are looking for the hate meaning as the stop words used in both context (hate, non-hate) and it never used by the hate community as a hate word. We normalized the textual data by lowercasing the words, and handle negation by converting "can't" to "can not". However, the misspelling is excluded from the data cleaning phase because most of the hate words (e.g., f*k) are misspelled or abbreviated in purpose to avoid detection. Stemming which removes prefix and suffix of the word, also excluded from the normalization phase to include some hate code words such as "blacks" that mostly refer to a race while the word "black" refers to a color.

Feature Extraction: The goal of this stage is to extract the domain-specific embedding vectors that represent the co-occurrence statistics of the word with its surrounding context. Genism provides word2vec library for this purpose, the parameters of the training model are the type of the model which is Continues Bag of Word (CBOW) because it performs slightly better than the Skip-Gram model (SG), with a window size=5 which represents the number of surrounding words, and vector size=300. The output of this stage is the embeddings for each word in the vocabulary to use it as a feature in the classification model. Classification: A deep sequential model structured by three layers is used for classification. The first layer is Bidirectional CuDNNLSTM, which is a fast LSTM implementation backed by a GPU-accelerated library of primitives for deep neural networks (CuDNN) (CUDA®, 2020; Abadi et al., 2020). Using BiLSTM maintains both forward and backward data of the input sentence. This makes the Bidirectional model to be familiar more with the context, but it consumes more computation time but we used CuDNNLSTM for accelerating the process on GPU. The second layer is a dense layer with a linear activation function chosen using grid search among other activation functions (relu, sigmoid, and none). The third layer is also a dense layer with a sigmoid activation function. The model compilation was performed with Binary Cross-Entropy Loss and optimization and Adam optimizer. The data is split into three categories: 60% training, 20% testing and 20% validation. The batch size is 256, and the training was done across 10 epochs.

- Approach 2: BERT Language Model

The second experiment was carried out using BERT, a pre-trained language model that had been fine-tuned for our objective. After a pre-trained model has been trained on a large generic text, fine-tuning is the process of training it on application-specific content. With the use of its embedding vectors, BERT encodes the input text. We used BERT for sequence classification model, which comprises a neural-network layer for classification.

The initial stages of BERT model converts the input sentence to tokens. The token embedding vector is created by adding the token, segment, and position embeddings together. For sentence classi-fication, BERT uses [CLS] short for classification, which is a unique token placed at the beginning of the sentence tokens to indicate the starting position of the classification task; in other words, the start-ing position of the fully connected layer to the last encoder layer, and finally to the softmax layer.

BERT released different versions that have dif-ferent properties based on the used language (e.g., Chinese, English, and Multilingual), the alphabet (i.e., Cased, Uncased), and the size of the layer structure (i.e., BERT-Base and BERT-Large). The BERT-Base model has 12 Transformer layers, each with 12 self-attention heads, and a total of 768 hidden states. The BERT-Large model has 24 trans-former layers, 16 self-attention heads, and a total of

1024 hidden layers. For training testing, the model parameters are LEARNING RATE = 2e 5, NUM TRAIN EPOCHS = 3:0, and BATCH SIZE = 16; 8 which are the parameter values recommended by the literature for sequence classification tasks.

## 5 Scope of the study

This study focuses on the detection of hate speech including its all types (e.g., race, sex, gender, etc.), and its levels including offensive language as two classes hate or not hate. The detection is based on the English text only.

## 6 Experiment and Results

This section includes a detailed description of the used datasets and technical details of each step in both approaches.

### 6.1 Datasets

We tested both approaches on three available datasets: Davidson-ICWSM (Davidson et al., 2017) dataset, Waseem-EMNLP (Waseem, 2016), and Waseem-NAACL (Waseem and Hovy, 2016) datasets and compare it with (Gupta and Waseem, 2017) results who used Hate Speech Word2Vec trained on 1 billion corpus size and LR classifier. The details of the datasets are shown in Table 1. Waseem and Hovy (2016) listed a number of crite-ria for identifying hate speech, including using a sexist or racial slur, attacking a minority, or promot-ing but not explicitly using hate speech or violent crime, among others, while Davidson et al. (2017) defined it as a language that is used to express ha-tred to a specific group or is meant to be derogatory, to humiliate, or to insult an individual of the group. Davidson et al. (2017) excluded the offensive lan-guage from their definition, the reason attributed is because offensive language is less hateful and more frequent use by the users, thus it should not be con-sidered as hate. However, According to Fortuna and Nunes (2018), hate speech could use offensive language but not necessary, which we agree with because frequent use of hate words does not mean that it should be socially acceptable, and no need to detect them and for this reason, we combined hate and offensive classes in their dataset to be hate class. Both of Waseem and Hovy (2016) and Davidson et al. (2017) definitions do not conflict with each other, and they agreed on the general hate speech definition. Thus, collapsing the labels and combining the datasets does not conflict with the general hate speech definition. Waseem's datasets have different classes, mainly racism, sexism, and neither and since we are handling the hate speech detection from neutral as to the best of our knowl-edge, there is no yet the state of the art solution for this problem as a binary classification task, and to compare it with Gupta and Waseem (2017), we collapsed the classes into two classes (e.g., racism and sexism as hate, neither as non-hate) because both of race and sex are types of hate speech. For Davidson's dataset, the offensive and hate speech are collapsed to be hate. Furthermore, we com-bined all the previously mentioned datasets in one dataset to assess the model performance on the largest possible diverse dataset that is balanced ac-cording to the lowest class number by randomly selecting a similar number of examples for each class and the number of classes specified according to the lowest class in the combined dataset. The dataset combining offers to us a hugely diverse set of data to assess the deep model performance on it as known that deep model performs better on large training data size.

| Dataset | Original labels | Number of hate classes | Number of non-hate classes | Total |
|---|---|---|---|---|
| Davidson-ICWSM | Hate speech, offensive, and neither | 20620 | 4163 | 24783 |
| Waseem-EMNLP | Racism, sexism, both, and neither | 1059 | 5850 | 6909 |
| Waseem-NAACL | Sexist, racist, and neither | 5406 | 11501 | 16910 |
| Balanced Combined Waseem, Waseem Hovey, and Davidson | Racism, sexism, and neither, both | 16260 | 16260 | 32520 |

Table 1: Datasets Description.

### 6.2 HSW2V and BiLSTM based Deep Model

For the first experiment, we investigated the perfor-mance of using Hate Speech Word2Vec (HSW2V) as features, and Bidirectional LSTM based deep model as a classifier. We compared our domain-specific embedding features (HSW2V) perfor-mance with domain agnostic embedding models, which are GloVe, and Google Word2Vec word em-beddings. We also compare results with domain-specific Hate Word2Vec (W2V-Hate) by Gupta and Waseem (2017) study. The details of each word embedding model are mentioned in the Table 2.

| Methods | Dimension | Trained on data of size | Pretrained on Platform |
|---|---|---|---|
| GoogleNews-vectors-negative[1] | 300 | 3 billion words | Google News |
| glove.6B.300d[2] | 300 | 6B tokens, 400K vocab | Wikipedia 2014 |
| glove.twitter.27B.200d[2] | 200 | 2B tweets, 27B tokens, 1.2M vocab | Twitter |
| W2V-Hate | 300 | 1 billion documents | Twitter |
| Hate speech Word2Vec (HSW2V) | 300 | 1,048,563 sentences, 116955 vocab, uncased | Twitter, hate websites |

Table 2: Details Description of Embedding Models.

The model performance is reported using weighted precision, recall, AUC, and f1-score to consider the class imbalance. The f1-score also is reported for each class separately to have a clear in-sight into the classifier performance on each class. The results are shown in Table 3. We also plotted the confusion matrix to show each of the FP, FN, TP, TN of the two used approaches.

As shown in Table 3, for each dataset, the maximum attained performance across different features is underlined, while the best performance among different classifiers is in BOLD. From a feature standpoint, we compare all of the embeddings for both domain agnostic and domain-specific embeddings using the same classifier (BiLSTM deep model). HSW2V, as shown in the table, outperforms all domain agnostic embedding models (Google Word2vec, GLoVe), considering the large range of corpus sizes (our corpus is 1M, other models are at least 2B), HSW2V could slightly outperforms domain agnostic approaches. The other variable that we consider in our comparison is the classification approach (LR by Gupta and Waseem (2017), and BiLSTM based deep model) and assess their performance with embedding features. The results show that the deep model surpasses the LR classifier. To show the deep model performance on a balanced dataset and to overcome the class imbalance influence on the deep model, we evaluate the proposed model on the combined dataset. This gives a decent sense of the performance, as well as the best result that our proposed approach can produce. The confusion matrix resulted from evaluating the model (Bidirectional LSTM and HSW2V) shown in Table 5. The deep model correctly classifies the most prevalent labels in the datasets.

[1] https://code.google.com/archive/p/ word2vec/

[2] https://nlp.stanford.edu/projects/ glove/

### 6.3 BERT Language Model

The BERT language model was used in the second experiment because it performs well across the board in NLP applications. BERT for sequence classification was implemented and fine-tuned us-ing datasets. Table 4 summarizes the findings of the testing evaluation.

| Methods | Datasets | P | R | f1-score (Hate) | f1-score (non-Hate) | f1-score | AUC |
|---|---|---|---|---|---|---|---|
| BERT Base | Davidson-ICWSM | 0.96 | 0.96 | 0.98 | 0.89 | 0.962 | 0.9309 |
| | Waseem-EMNLP | 0.92 | 0.92 | 0.7654 | 0.9541 | 0.9216 | 0.8455 |
| | Waseem-NAACL | 0.85 | 0.85 | 0.7612 | 0.8881 | 0.8472 | 0.8227 |
| | Combined Balanced | 0.95 | 0.95 | 0.9543 | 0.9552 | 0.9547 | 0.9547 |
| BERT Large | Davidson-ICWSM | 0.96 | 0.96 | 0.9788 | 0.8924 | 0.9646 | 0.9345 |
| | Waseem-EMNLP | 0.91 | 0.91 | 0.6939 | 0.9458 | 0.9103 | 0.8371 |
| | Waseem-NAACL | 0.85 | 0.85 | 0.7643 | 0.8937 | 0.8521 | 0.823 |
| | Combined Balanced | 0.96 | 0.96 | 0.962 | 0.9625 | 0.9623 | 0.9623 |

Table 4: BERT for sequence classification hate speech experiment results (Base-Large)

The performance of the BERT classifier was highly acceptable and desirable. In addition, for datasets of greater size, BERT Large surpasses BERT Base with a pretty similar results to that of BERT Base. Due to the computational requirements for BERT large, most current research avoids utilizing it. However, the overall performance of BERT overcomes all of embedding models with the deep model that we proposed as first approach.

## 7 Discussion

For the feature extraction stage, we evaluate the influence of using domain-specific word embed-ding of hate speech, which helps the model expose the most used terms, abbreviations, and intentional spelling mistakes by the community who post in a given domain (Badjatiya et al., 2017). This is the main reason to investigate the domain-specific embedding with deep models. We applied word similarity which finds the closest word to the in-put word according to the cosine distance between them. Table 6 shows the result of applying word similarity on intentionally misspelled word that is commonly used by the hate community (fc*). As shown in the table, domain-agnostic embedding models failed to retrieve similar words, while our HSW2V was able to retrieve other intentionally

| Source | Classifier | Methods | Dataset | P | R | f1-score (hate) | f1-score (no-hate) | f1-score | AUC |
|---|---|---|---|---|---|---|---|---|---|
| Gupta and Waseem (2018) | Logistic Regression (LR) | Hate W2V(300) | Davidson ICWSM | 0.91 | 0.91 | - | - | 0.912 | 0.84 |
| | | | Waseem EMNLP | 0.84 | 0.86 | - | - | 0.844 | 0.638 |
| | | | Waseem NAACL | 0.76 | 0.77 | - | - | 0.75 | 0.679 |
| The proposed methods | Bidirectional LSTM deep model | GoogleNews-vectors-negative300 | Davidson ICWSM | 0.9718 | 0.964 | 0.9679 | 0.8457 | 0.9473 | 0.9132 |
| | | | Waseem EMNLP | 0.8466 | 0.5594 | 0.6737 | 0.9484 | 0.9033 | 0.7697 |
| | | | Waseem NAACL | 0.6899 | 0.6713 | 0.6805 | 0.8542 | 0.799 | 0.7654 |
| | | | Combined balanced | 0.9407 | 0.9323 | 0.9365 | 0.9376 | 0.9371 | 0.937 |
| | | GloVe.6B.300d | Davidson ICWSM | 0.9705 | 0.9587 | 0.9646 | 0.831 | 0.9421 | 0.9073 |
| | | | Waseem EMNLP | 0.75 | 0.5943 | 0.6631 | 0.9463 | 0.9028 | 0.7792 |
| | | | Waseem NAACL | 0.7047 | 0.6558 | 0.6794 | 0.8569 | 0.8002 | 0.7633 |
| | | | Combined balanced | 0.9404 | 0.9424 | 0.9414 | 0.9413 | 0.9414 | 0.9414 |
| | | GloVe.Twitter.27B.200d | Davidson ICWSM | 0.9625 | 0.9728 | 0.9676 | 0.8347 | 0.9453 | 0.8927 |
| | | | Waseem EMNLP | 0.6527 | 0.7358 | 0.6917 | 0.9399 | 0.9018 | 0.8324 |
| | | | Waseem NAACL | 0.677 | 0.7058 | 0.6911 | 0.8503 | 0.7994 | 0.7738 |
| | | | Combined balanced | 0.9484 | 0.928 | 0.9381 | 0.9394 | 0.9387 | 0.9388 |
| | | HSW2V(300) | Davidson ICWSM | 0.9724 | 0.9682 | 0.9703 | 0.8551 | <u>0.9509</u> | 0.9162 |
| | | | Waseem EMNLP | 0.723 | 0.665 | 0.6928 | 0.9469 | <u>0.9079</u> | 0.8094 |
| | | | Waseem NAACL | 0.7118 | 0.6993 | 0.7055 | 0.8634 | <u>0.8129</u> | 0.7831 |
| | | | Combined balanced | 0.9517 | 0.9341 | 0.9428 | 0.9439 | <u>0.9434</u> | 0.9434 |

Table 3: Results of Bidirectional LSTM based deep model on the datasets.

misspelled words that close in the meaning to the input word.

| The word | Word2Vec | Glove | HSW2V |
|---|---|---|---|
| fc* (missplled) | hahah,lmfao, questlove @, u,ppl,hahahahahaha, JeremyShockey @, ummmmm, ROTFLMAO, Awright, freaken | h?ndbold, nordsj?lland, k?benhavn,br?ndby, hik, parken, s?nderjyske, genelerbirli?i,¯ 1972/73, iwr, al-hilal | fu*, lmfaoo, shii, fucc, dese , lmaooooo, dats, noooo, lmfaoooo, ay |

Table 6: Word similarity of misspelled hate word fc*

The results of the first experiment of using the first approach (Table 3) showed that using a domain-specific embedding model (HSW2V) was very competitive to the domain agnostic embed-ding models (Word2Vec, GloVe) although there is a huge difference between the corpus size, it is 1M for HSW2V and at least 2B for domain agnostic embedding models taking into consideration that the classifier is the same (BiLSTM deep model), which confirmed that domain-specific word em-beddings outperform domain-agnostic word em-bedding models, because it is more knowledgeable about the hate domain, while domain-agnostic are trained on books and Wikipedia which rarely have hate community context.

From the classifier perspective, we compared our BiLSTM deep model with LR experiment by Gupta and Waseem (2017), the results showed that the BiLSTM based deep model outperforms LR classifier, the BiLSTM based deep model improves the performance with at least 5 percent.

In the second approach, we used BERT model on the hate speech binary classification task. Ta-ble 4, report the result of experiments on both Base and Large models. Because the BERT model is deeply bidirectional and trained on huge data sets, it outperforms all other distributional based embeddings, including domain-agnostic (e.g., Google Word2Vec, GloVe) and domain-specific (e.g., HSW2V). BERT also has an intuitive training procedure for its vocabulary as it includes sub-words instead of complete words. Although the simple training procedure of domain-specific embeddings, the performance was not too low in comparison with BERT. Domain-specific embeddings overcome BERT model embedding in that it includes intentionally misspellings and commonly hate words that BERT fails to retrieve when we searched about specific words in Bert Base vocabu-lary as shown in Figure 1, and this is because BERT also trained on books and Wikipedia which rarely includes these words.

| Datasets/model | Davidson-ICWSM | Waseem-EMNLP | Waseem-NAACL | Balanced Dataset |
|---|---|---|---|---|
| HSW2V(300), Bidirectional LSTM deep model | not hate: 754 / 129; hate: 171 / 3,953; accuracy=0.9395, misclass=0.0605 | not hate: 1,132 / 38; hate: 86 / 126; accuracy=0.9103, misclass=0.0897 | not hate: 2,021 / 280; hate: 292 / 789; accuracy=0.8309, misclass=0.1691 | not hate: 3,086 / 166; hate: 256 / 2,996; accuracy=0.9351, misclass=0.0649 |
| BERT Large | non hate: 767 / 71; hate: 81 / 4,037; accuracy=0.9693, misclass=0.0307 | non hate: 1,092 / 72; hate: 53 / 164; accuracy=0.9095, misclass=0.0905 | non hate: 2,057 / 220; hate: 207 / 897; accuracy=0.8737, misclass=0.1263 | non hate: 3,181 / 110; hate: 122 / 3,090; accuracy=0.9643, misclass=0.0357 |

Table 5: Confusion Matrix of HSW2V and BERT.

Figure 1: BERT Base vocabulary search for misspelling and hate term

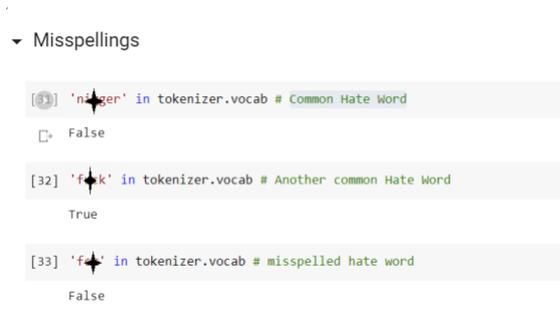

Finally, Table 5 shows the confusion matrix of HSW2V(300), the Bidirectional LSTM deep model compared with BERT language model to assess their performance on each dataset and graphically represents the performance of each model.

## 8 BERT interpretation using LIME

To gain more knowledge about BERT model performance in the classification task of this study, we applied LIME (Ribeiro et al., 2016). LIME stands for Local Interpretable Model-agnostic Explana-tions, a strategy for understanding the model by modifying data samples and seeing how the pre-dictions change by looking at internal properties and how they connect to specific predictions. We employed LIME on Waseem-EMNLP dataset and BERT base model. We reported different cases as follow:

- Case 1 : True Positive

LIME highlights the words contributed more to classify the sentence with a different color for each class, Figure 2 shows the case in which the sentence actual and predicted class is hate, and which words contribute in classifying this sentence as hate. The color intensity increased according to the word more contributed to the prediction such as "n*gga".

Figure 2: Result of applying LIME on document actual label=1 and predicted=1

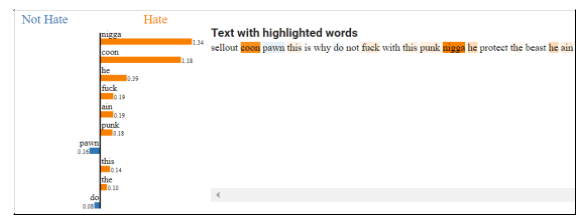

- Case 2 : True Negative

Figure 3 shows the case in which the sentence ac-tual and predicted class is not hate, and which words contribute to classify this sentence as not hate with darker blue color such as "announcement", while the words with orange color contributed more to classify the sentence to be hate, however, in this case, the weight of the blue-colored words is more than the orange-colored words, thus the classifier predict this sentence as not hate which agree with human sense.

Figure 3: Result of applying LIME on document actual label=0 and predicted=0

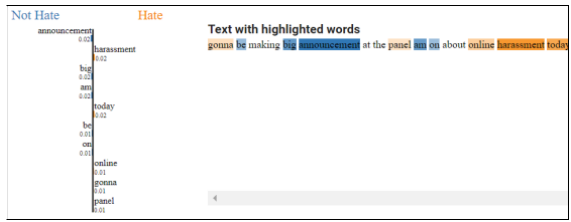

- Case 3 : False Positive

We also apply LIME to analyze error classifications. Figure 4 shows the case in which the sentence actual class is not hate but predicted class is hate, and which words contribute in classifying this sentence as hate which is "feminazi" word that influences on the classifier to predict it as hate.

Figure 4: Result of applying LIME on document actual label=0 and predicted=1

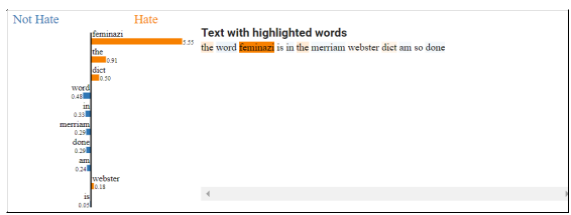

- Case 4 : False Negative

Figure 5 shows the case in which the sentence actual class is hate but predicted class is not hate, and which words contribute in classifying this sentence as not hate. It seems in this situation, the classifier is more accurate than the human annotator as this sentence is not a hate sentence as it is obvious from the observed intent of the context writer.

Figure 5: Result of applying LIME on document actual label=1 and predicted=0

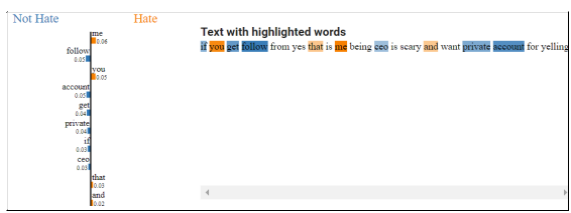

## 9 Conclusion and Future Work

To conclude, BERT design provides an appropriate feature extraction and classification procedure for hate speech detection. BERT combines the benefits of domain agnostic and domain-specific word embedding by train the model on vast data then add an extra layer to trained on domain-specific data (fine-tuning). BERT also saves effort and time for building an embedding model from scratch. However, domain-specific word embedding overcomes BERT model in that it can detect hate terms and abbreviations and intentionally misspellings meaning. This study can be extended to detect multi-class hate speech for future work.